\documentclass[journal,twoside,web]{ieeecolor}
\usepackage{generic}
\usepackage{cite}
\usepackage{amsmath,amssymb,amsfonts}
\usepackage{algorithmic}
\usepackage{graphicx}
\usepackage{algorithm,algorithmic}
\usepackage{hyperref}
\hypersetup{hidelinks=true}
\usepackage{textcomp}
\usepackage{subcaption}
\usepackage{multirow}
\usepackage{booktabs}
\def\BibTeX{{\rm B\kern-.05em{\sc i\kern-.025em b}\kern-.08em
    T\kern-.1667em\lower.7ex\hbox{E}\kern-.125emX}}
\begin{document}
\title{An Enhanced Pyramid Feature Network Based on Long-Range Dependencies for Multi-Organ Medical Image Segmentation}
\author{Dayu Tan, Cheng Kong, Yansen Su, Hai Chen, Dongliang Yang, Junfeng Xia, and Chunhou Zheng
\thanks{This work was supported in part by the National Key Research and Development Program of China (2021YFE0102100), in part by National Natural Science Foundation of China (62303014, 62172002, 62322301). (\emph{Corresponding author: Yansen Su.})}
\thanks{Dayu Tan, Chen Kong, Yansen Su, Junfeng Xia, and Chunhou Zheng are with the Key Laboratory of Intelligent Computing and Signal Processing, Ministry of Education, Anhui University, Hefei 230601, China (e-mail: suyansen@ahu.edu.cn).}
\thanks{Hai Chen and Dongliang Yang are with the Second Department of Thoracic Surgery, Anhui Chest Hospital, Hefei 230022, China (e-mail: chanhai@aliyun.com, joyyoung2269@163.com).}}

\maketitle

\begin{abstract}
In the field of multi-organ medical image segmentation, recent methods frequently employ Transformers to capture long-range dependencies from image features. However, these methods overlook the high computational cost of Transformers and their deficiencies in extracting local detailed information. To address high computational costs and inadequate local detail information, we reassess the design of feature extraction modules and propose a new deep-learning network called LamFormer for fine-grained segmentation tasks across multiple organs. LamFormer is a novel U-shaped network that employs Linear Attention Mamba (LAM) in an enhanced pyramid encoder to capture multi-scale long-range dependencies. We construct the Parallel Hierarchical Feature Aggregation (PHFA) module to aggregate features from different layers of the encoder, narrowing the semantic gap among features while filtering information. Finally, we design the Reduced Transformer (RT), which utilizes a distinct computational approach to globally model up-sampled features. RRT enhances the extraction of detailed local information and improves the network's capability to capture long-range dependencies. LamFormer outperforms existing segmentation methods on seven complex and diverse datasets, demonstrating exceptional performance. Moreover, the proposed network achieves a balance between model performance and model complexity. The code has been made available on GitHub:
\url{https://github.com/kec1212/LamFormer}.
\end{abstract}


\section{Introduction}
\label{sec:introduction}
\IEEEPARstart{M}{edical} image segmentation is a cornerstone of the medical imaging field, providing indispensable support for clinical decision-making, treatment planning, and disease monitoring. Accurate segmentation plays a crucial role in medical image processing, as it involves identifying regions of interest within medical images, extracting relevant features and providing reliable information to assist doctors in diagnosis. Multi-organ segmentation\cite{b1} is essential in medical image analysis. Given the significant time and labour costs associated with manual medical imaging and the meticulous nature of medical science, segmentation methods are vital for achieving accurate results and efficient processes\cite{b2}. The shape and size of different organs in multi organ medical images are complex and changeable, which leads to poor recognition of small sized organs by the segmentation network. In the context of medical image segmentation, certain pathological regions often occupy only a small fraction of the image, while normal tissues dominate the majority of the area. This inbalance in segmentation classes severely impacts the performance of segmentation networks. Both long-range dependencies and local details are critical for segmentation outcomes\cite{b3}. However, effectively integrating long-range dependency modeling with detailed information extraction remains a significant challenge in segmentation networks. 
\par With the emergence of deep learning (DL), DL-based methods for medical image segmentation have been proposed to segment multi-organ regions automatically. The Fully Convolutional Network (FCN)\cite{b4} achieves pixel-level semantic segmentation of images by replacing the fully connected layers in traditional Convolutional Neural Networks (CNNs)\cite{b5} with convolutional layers. U-Net\cite{b6} is famous for its simple and effective U-shaped encoder-decoder\cite{b7} architecture. CNNs increasingly develop into powerful and complex architectures. Furthermore, Transformer\cite{b8} is initially designed for sequence modeling and transduction tasks in natural language processing (NLP)\cite{b9}. Since the self-attention mechanism in Transformer can capture long-range dependencies, many researchers apply it to medical image segmentation. Recently, Mamba has been known for its efficient computational performance and ability to flexibly handle long sequences, and it is considered a strong alternative to Transformers. Mamba has been applied in the computer vision domain to address tasks such as object detection, segmentation, and classification.
\par In the multi-organ medical image segmentation\cite{b10}, the significant differences in size and shape of each organ often lead to poor segmentation outcomes by the networks. CNNs are affected by fixed receptive fields, resulting in inadequate segmentation of large-sized organs. In contrast, Transformers excel global modeling but cannot extract detailed information, resulting in insufficient segmentation of small organs. Furthermore, the self-attention mechanism\cite{b36} has quadratic computational complexity, which requires substantial computational resources and results in slow processing speeds. Mamba achieving linear computational complexity while capturing long-range dependencies\cite{b35}, which improves the processing efficiency of long sequences and is conducive to the application in high-resolution image segmentation tasks.
\par In this study, we introduce a novel deep-learning network named LamFormer, which addresses the limitations of both CNNs and Transformers. LamFormer incorporates Linear Attention Mamba (LAM), Parallel Hierarchical Feature Aggregation (PHFA), and Reduced Transformer (RT) modules. The LAM enhances nonlinear feature expression and captures long-range dependencies. The PHFA integrates features from various levels, narrowing their semantic gaps. The RT further captures long-range dependencies among the upsampling features and minimizes the loss of spatial information due to encoder downsampling. The main contributions of this work are as follows:
\begin{itemize}
\item We design a novel feature extraction module called Linear Attention Mamba. This module captures long-range dependencies with linear complexity, thus improving the network’s capacity to represent low- and high-dimensional features. Furthermore, the Linear Attention Mamba complements extracting local detail information.

\item We propose the Parallel Hierarchical Feature Aggregation module to effectively fuse features from the downsampling stages. This module significantly reduces semantic gaps between low- and high-dimensional features and filters redundant information in different hierarchical features.

\item This study presents a novel segmentation network that utilizes a feature pyramid structure to extract multi-scale long-range dependencies from different hierarchical features. The proposed network exhibits strong robustness and generalization capability for complex and diverse datasets.
\end{itemize}

\section{Related Work}
\subsection{Medical Image Segmentation}
Medical image segmentation is a pixel-level task to isolate lesions or organs within a given image. Recent advances in deep-learning\cite{b12} have replaced traditional neural network methods due to limitations such as fixed input dimensions, low segmentation efficiency, and high storage costs. FCN addresses these issues, achieving pixel-level classification without input size constraints, improving applicability, and reducing loss of detail. U-Net has evolved into a widespread reference network with its compact architecture and reliable performance. Subsequent attention mechanisms further enhance segmentation network performance, as demonstrated by Attention-UNet, which uses gated units to suppress irrelevant information and highlight local critical features. The Convolutional Block Attention Module (CBAM)\cite{b13} calculates attention at both channel and spatial levels, improving perceptual capabilities without increasing network complexity.

\subsection{Long-Range Dependencies Modeling}
In recent years, Transformer has emerged as a core architecture in computer vision and natural language processing, demonstrating impressive performance in various tasks. The latest advancements in Transformer networks and their variants significantly improve the modeling of long-range dependencies in computer vision. DaViT\cite{b14} employs a dual attention mechanism across spatial and channel dimensions to efficiently achieve global modeling. LongNet\cite{b15} introduces an expanded attention mechanism, where the attention range grows exponentially with distance. This expanded attention can be seamlessly integrated into Transformer blocks, replacing standard attention mechanisms. Longformer\cite{b16} combines local window attention with global attention to establish local context representations and complete sequence representations for predictions. Although DeViT\cite{b17} features a complex structure, it employs decomposed self-attention to capture long-range dependencies at the cost of local windows.
\par Mamba emerges as a novel deep-learning architecture, showcasing exceptional performance in handling long-sequence data, particularly demonstrating a clear advantage in competition with Transformer networks. Vision Mamba\cite{b18} utilizes bidirectional Mamba modules for bidirectional sequence modeling tailored to visual tasks. MLLA\cite{b19} reveals striking similarities between the Mamba and linear attention, elucidating the critical factors behind Mamba’s success. MSVMamba\cite{b20} processes feature maps using multi-scale 2D scanning techniques, effectively capturing long-range dependencies while reducing computational costs. Altogether, related networks are dedicated to employing a single long-range dependencies modeling approach for extracting features without considering diverse methods for global modeling.

\begin{figure*}[h]
    \centering
    \includegraphics[width=0.98\textwidth]{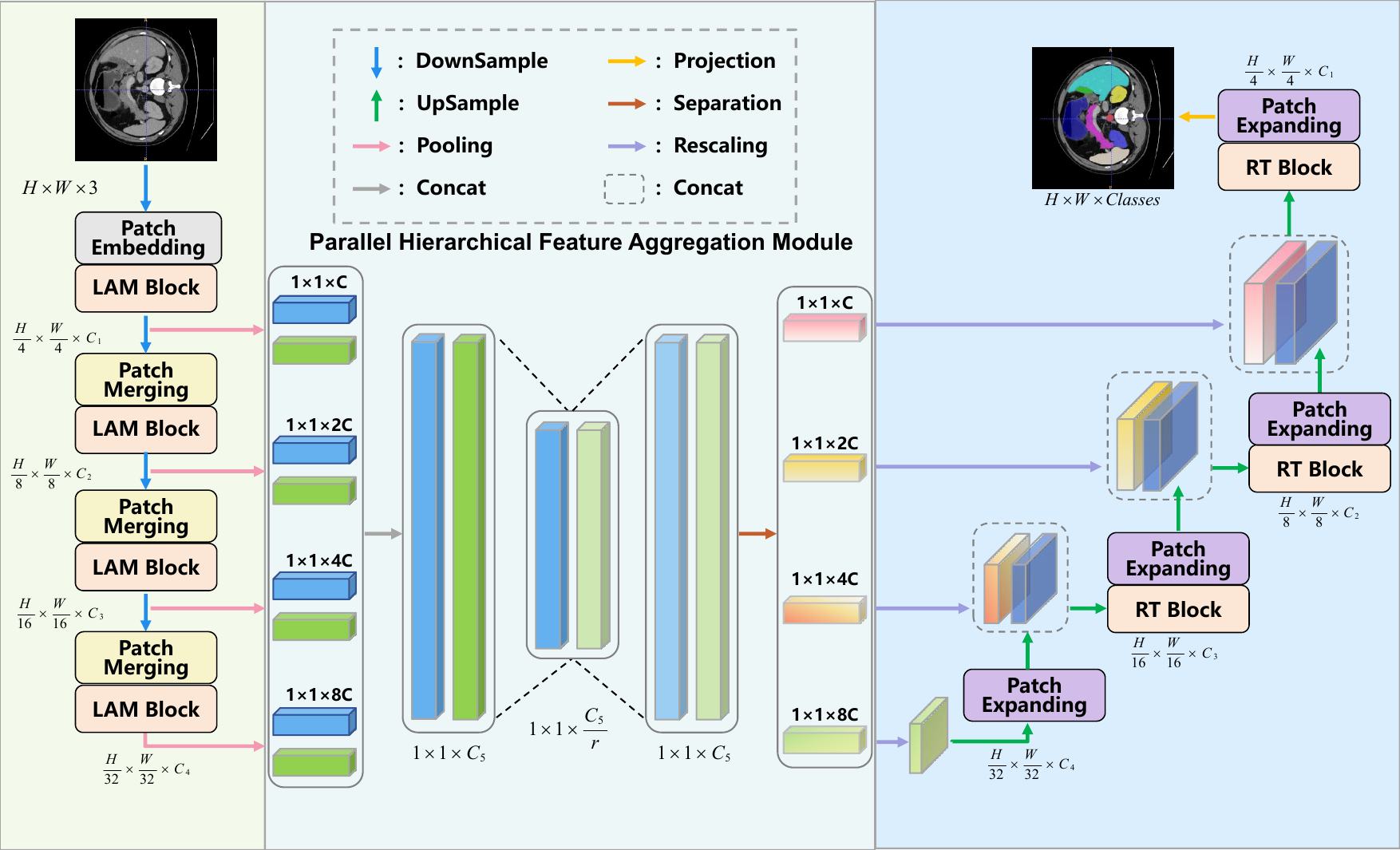}
    \caption{Illustration of the proposed LamFormer. We utilize the Linear Attention Mamba (LAM) to construct the encoder component, employ Parallel Hierarchical Feature Aggregation (PHFA) to replace the traditional skip connections, and implement the Reduced Transformer (RT) in the up-sampling section.}
\end{figure*}

\subsection{U-shaped Structure for Medical Image Segmentation}
The U-shaped architecture originates from U-Net, which consists of an encoder responsible for downsampling, a decoder for upsampling, and simple skip connections. Even though U-Net lacks complex modules, it achieves impressive performance due to its effective U-shaped configuration\cite{b21}. Subsequently, many medical image segmentation networks adopt the U-shaped structure. UNet++\cite{b22} builds upon the U-shaped architecture of U-Net by incorporating dense connections in the skip connection, preserving detailed information from each layer and achieving improved segmentation performance. Similarly, Attention-UNet retains the U-shaped structure while integrating attention mechanisms into the skip connections, emphasizing relevant regions and suppressing less critical areas.
\par Transformer-based networks also widely embrace the U-shaped architecture. For instance, TransUNet\cite{b23} combines CNN with Transformer by introducing Transformer blocks at the lower levels of the encoder, thereby enhancing segmentation accuracy. Swin-UNet\cite{b11} employs Swin Transformer\cite{b24} blocks as its backbone, forming a U-shaped network based purely on the Transformer. Nowadays, Mamba is also applied to medical image segmentation. VM-Unet\cite{b26} introduces Visual State Space (VSS) blocks to create a U-shaped network based on pure state space models (SSMs). Swin-UMamba\cite{b27} leverages the advantages of pretraining to enhance the performance of Mamba-based networks. U-Mamba\cite{b28} designs a hybrid CNN-SSM block that combines the local feature extraction of convolutional operations with the long-range dependencies modeling of SSMs. In summary, the U-shaped structure\cite{b29} develops a classic segmentation architecture, exerting a profound and positive impact on medical image segmentation.

\section{Proposed Method}
This section introduces LamFormer's architecture and describes how its key components are implemented. As shown in Fig. 1, LamFormer consists of three main components: an encoder, a decoder, and skip connections. The designed network employs a four-layer feature pyramid structure. The resolution of the input image decreases progressively across layers while the channel dimension increases, facilitating the extraction of features at different resolutions. The resolutions at each layer are denoted as \(\frac{H}{4}\times\frac{W}{4}\), \(\frac{H}{8}\times\frac{W}{8}\), \(\frac{H}{16}\times\frac{W}{16}\), and \(\frac{H}{32}\times\frac{W}{32}\), where \(H\) and \(W\) represent height and width, respectively.

\subsection{Linear Attention Mamba in Encoder}
We propose a module based on linear attention, referred to as Linear Attention Mamba (LAM), as illustrated in Fig. 2. LAM consists of Layer Normalization (LN), Long-Range Linear Attention (LRLA), and Forward Residual Network (FRN) components.

\par The Long-Range Linear Attention (LRLA), illustrated on the right side of Fig. 2, employs the Mamba structure. The input is processed through a linear layer and split into two branches. The main branch undergoes processing via a depthwise separable convolution and an activation function before computing the attention weights within the linear attention. Then, the output from the main branch is combined with the secondary branch through element-wise multiplication to merge the two branches. Finally, a linear layer is utilized to mix the features. The computational method for Long-Range Linear Attention is as follows:
\begin{equation}
LRLA(X)=W_2(LA(\sigma(DConv(W_1x))))\cdot\sigma(W_1x),
\end{equation}
where \(x \in \mathbb{R}^{N \times C}, N=H \times W\) represents the features after Layer Normalization. \(W_1\) and \(W_2\) are learnable weight matrices used for linear transformations of the features. \(DConv(\cdot)\) denotes depthwise separable convolution, while \(\sigma\) refers to the \(SiLU\) activation function. \(LA\) represents linear attention.
\par In LRLA, linear attention is utilized to compute the attention weights. Unlike the widely used Softmax attention, which relies on the quadratic computational complexity of the dot-product operation, linear attention replaces the nonlinear Softmax operation with linear normalization. The computation formula for linear attention is as follows:
\begin{equation}
Q=\phi(xW_Q), K=\phi(xW_K), V=xW_V,
\end{equation}
\begin{equation}
Z_i=\sum_{j=1}^{N} \frac{Q_i K_j^T}{\sum_{j=1}^{N} Q_i K_j^T} V_j=
\frac{\sum_{j=1}^{N} Q_i (K_j^T V_j)}{\sum_{j=1}^{N} Q_i K_j^T},
\end{equation}
in Eqs. (2) and (3), \(x\in\mathbb{R}^{N \times C}\) denotes input features, \(\phi\) denotes kernel function. \(Q\in\mathbb{R}^{N \times d}, K\in\mathbb{R}^{N \times d}, V\in\mathbb{R}^{N \times C}\) represent the \(Query\), \(Key\), and \(Value\) matrices. \(W_Q,W_K\in\mathbb{R}^{C \times d}, W_V\in\mathbb{R}^{C \times C}\) denote the projection matrix, and \(Q_i, K_j\in \mathbb{R}^{1 \times d}, K_j\in\mathbb{R}^{1 \times C}\) represent an individual token. By utilizing the associative property of matrix multiplication, we can change the computation order from \((QK^T)V\) to \(Q(K^T V)\). The change of computation order produces a linear computational complexity of \(O(N)\). Consequently, linear attention enables global modeling for longer sequences.
\par The structure of Forward Residual Network (FRN), as illustrated on the left side of Fig. 2. In order to further extract the local details in the image, a convolution operation is embedded in the traditional Forward Feedback Network(MLP). Due to the embedding of convolutions leads to gaps among features, layer normalization is employed to align the features and their distributions. Furthermore, the residual connection is used to promote the propagation of features among different normalization layers and to enhance the ability to extract local details. The computation method for the FRN is as follows:
\begin{equation}
\begin{aligned}
f_1 &= LN_1(DConv(FC_1(x)) + FC_1(x)), \\
f_2 &= LN_2(f_1 + FC_1(x)), \\
f_3 &= FC_2(GELU(LN_3(f_2 + FC_1(x)))),
\end{aligned}
\end{equation}
where \(DConv(\cdot)\) denotes depthwise separable convolution, which can reduce the number of parameters in the convolution operations and enhance processing efficiency. FC represents full connection layer, and LN is layer normalization.

\begin{figure}
    \centering
    \includegraphics[width=0.47\textwidth]{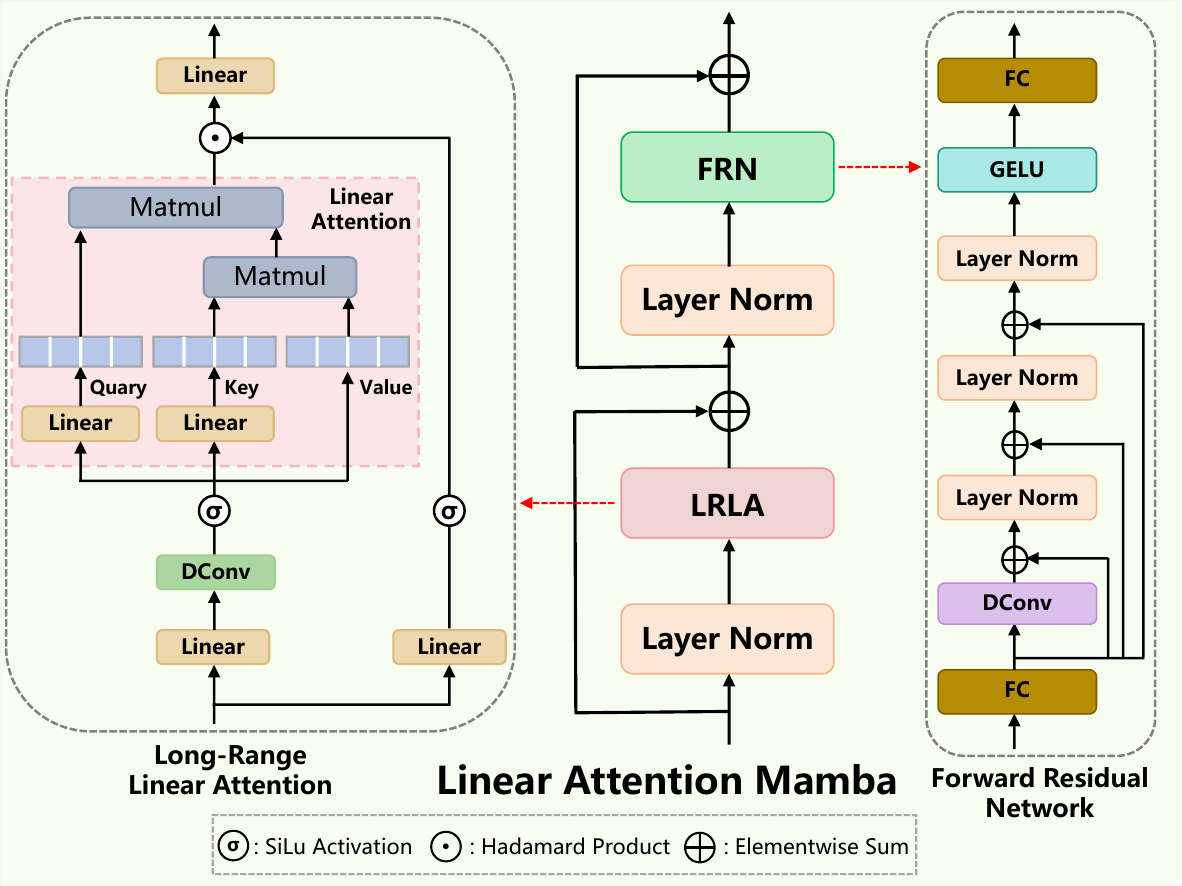}
    \caption{The Linear Attention Mamba (LAM) block is constituted by Layer Normalization (LN), Long-Range Linear Attention (LRLA), and the Forward Residual Network (FRN).}
\end{figure}

\subsection{Parallel Hierarchical Feature Aggregation in Skip Connection}
To integrate multi-scale features from different stages and filter redundant information, we propose the Parallel Hierarchical Feature Aggregation (PHFA) module, which constitutes the skip connection part of the network. The PHFA module is depicted in Fig. 3.

\par Mathematically, we analyze the input to the PHFA module. We consider \(F_i\in\mathbb{R}^{C \times H \times W}\), where \(F_i\) represents the features from the \(i-th\) layer of the encoder. Within the PHFA module, Global Max Pooling (GMP) and Global Average Pooling (GAP) are used to compress the spatial dimensions, resulting in vectors with \(k\) channels, denoted as \(f(x)\) and \(g(x)\).
\begin{equation}
f(x) = Max\{X^k (i,j)\}, 1 \leq i \leq H, 1 \leq j \leq W,
\end{equation}
\begin{equation}
g(x) = \frac{1}{H \times W} \sum_{i=1}^H \sum_{j=1}^W X^k (i,j),
\end{equation}
in Eqs. (5) and (6), \(f(x), g(x) \in \mathbb{R}^{C \times 1 \times 1}\).
\par Subsequently, the resulting two sets of short vectors (each set comprising four vectors of different lengths) are concatenated individually and fed into a shared fully connected layer for information interaction.
\begin{equation}
\begin{aligned}
V_1 &= Concat(f(E_i)), \\
V_2 &= Concat(g(E_i)), \\
V_3 &= \sigma (MLP(V_1) + MLP(V_2)),
\end{aligned}
\end{equation}
where \(V_1, V_2, V_3 \in \mathbb{R}^{\widetilde{C} \times 1 \times 1}\)denote feature vectors, \(Concat(\cdot)\) represents the Concatenation operation. \(\sigma\) denotes \(Sigmoid\) activation function, \(MLP(\cdot)\) represents shared fully connected layer.

\par Finally, the two long vectors containing weighted information are summed and combined into a single vector, which is then passed through an activation function. The resulting vector is separated into four vectors of different lengths and multiplied channel-wise with the input features. Therefore, feature maps with integrated multi-scale semantic information are obtained.

\begin{figure*}[h]
    \centering
    \includegraphics[width=0.97\textwidth]{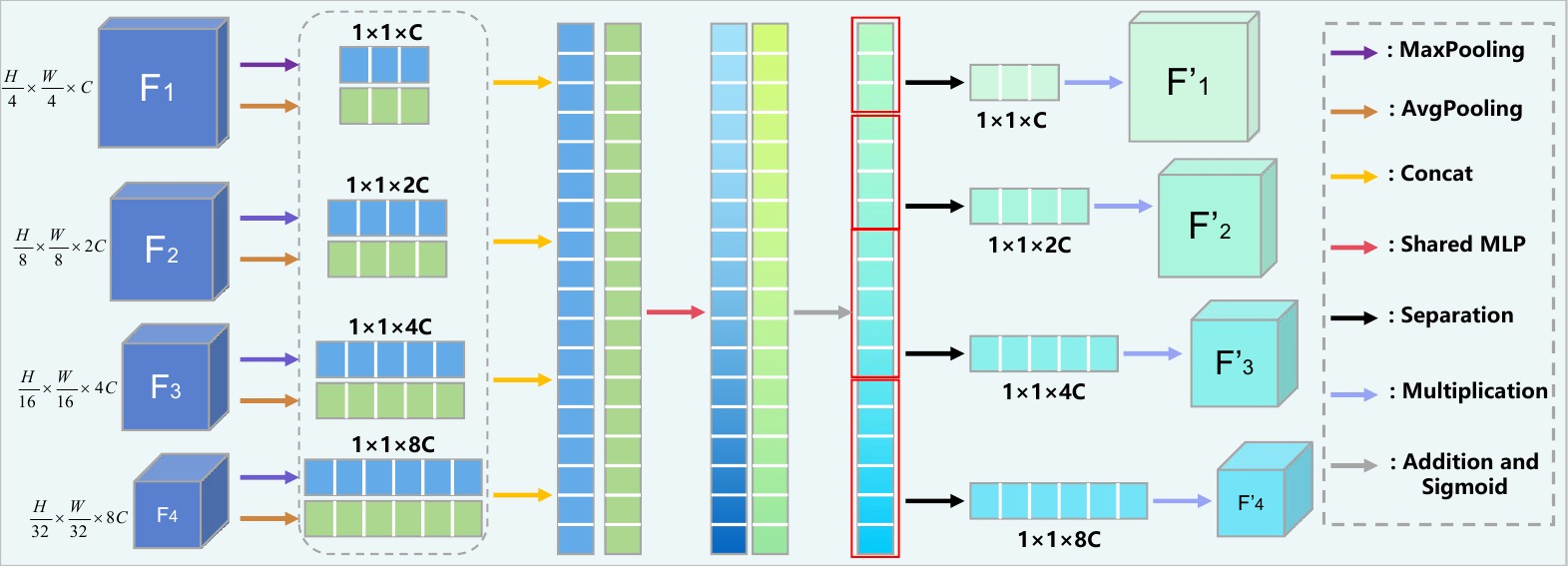}
    \caption{The Parallel Hierarchical Feature Aggregation (PHFA) module is implemented through parallel pooling operations, multilayer perceptrons (MLP) with shared weights and separation operations.}
\end{figure*}
\begin{figure}[h]
    \centering
    \includegraphics[width=0.47\textwidth]{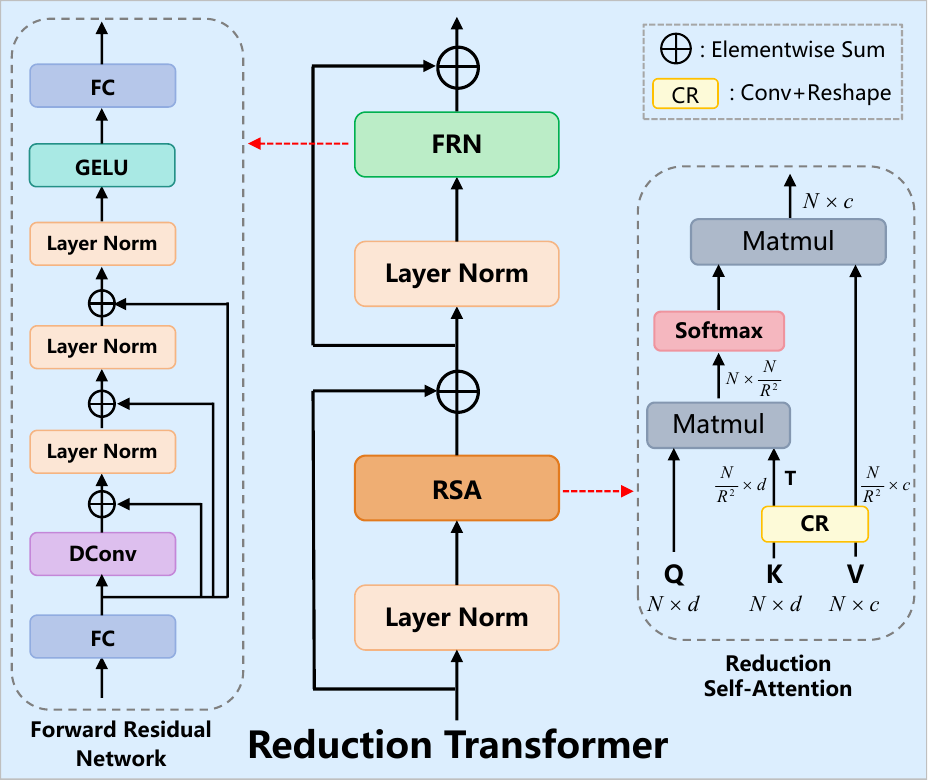}
    \caption{The Reduced Transformer (RT) block consists of Layer Normalization (LN), Reduced Self-Attention (RSA), and Forward Residual Network (FRN).}
\end{figure}

\subsection{Reduced Transformer in Decoder}
The decoder consists of Reduced Transformer (RT) blocks and Patch Expanding layers, with the latter responsible for the upsampling operations to restore the spatial scale of the feature maps. Mamba and Transformer both excel at capturing long-range dependencies, but they employ different methodologies. We further integrate the RT block to enable global feature modeling during upsampling. Combining linear attention and self-attention mechanisms can enhance the ability to capture long-range dependencies.
\par Due to the quadratic computational cost associated with processing sequence lengths, Transformer is less suited for handling high-resolution image data. The computation method for self-attention in Transformer is as follows:
\begin{equation}
y = Softmax(\frac{QK^T}{\sqrt{d_k}}) \cdot V,
\end{equation}
in Eq. (8), \(Q, K \in \mathbb{R}^{N \times d}\), and \(V \in \mathbb{R}^{N \times C}\) represent the \(Query\), \(Key\), and \(Value\) matrices, respectively. The Softmax function computes the similarity between each \(Query\) and \(Key\) pair, resulting in a computational complexity of \(O(N^2)\). High computational cost leads to an unbearable consumption of computational resources when dealing with high-resolution image data.
\par To solve the problem of calculating costs, we employ convolution operations to decrease the size of the \(K\) and \(V\) matrices, thereby reducing computational costs and facilitating the processing of high-resolution image data.
\begin{equation}
\widetilde{y} = Softmax(\frac{Q \widetilde{K}^T}{\sqrt{d_k}}) \cdot \widetilde{V},
\end{equation}
\begin{equation}
\begin{aligned}
\widetilde{K} &= LN(reshape(Conv(x))W_K), \\
\widetilde{V} &= LN(reshape(Conv(x))W_V),
\end{aligned}
\end{equation}
where \(x\in\mathbb{R}^{C \times H \times W}\) represents input, \(Conv(\cdot)\) denotes the convolution operation used to reduce the input. After convolution, the shape of \(x\) changes to \(C \times \frac{H}{R} \times \frac{W}{R}\), which is then reshaped using the reshape operation to obtain a new shape of \(\frac{HW}{R^2} \times C\). The projection matrices are represented as \(W_K \in \mathbb{R}^{C \times d}, W_V \in \mathbb{R}^{C \times C}\), and LN denotes layer normalization. \(\widetilde{K} \in \mathbb{R}^{\frac{N}{R^2} \times d}, \widetilde{V} \in \mathbb{R}^{\frac{N}{R^2} \times C}, N = H \times W\) represent the \(Key\) and \(Value\) matrices after spatial reduction. Therefore, the computational complexity of the RSA is \(O(\frac{N^2}{R^2})\).

\section{Experiments}
We conduct experiments on the following seven datasets. Aside from the SLIVER and LungCancer datasets, the remaining five are multi-organ segmentation datasets.

\begin{figure*}[h]
    \centering
    \begin{subfigure}{0.48\linewidth}
        \centering
        \includegraphics[width=0.98\linewidth]{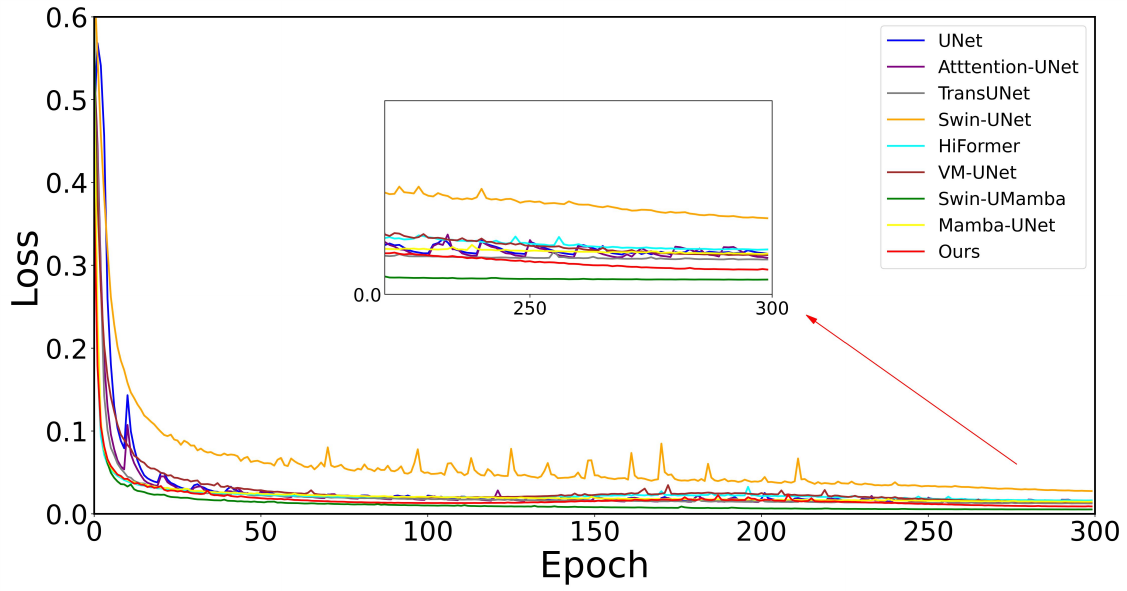}
        \caption{MM-WHS}
    \end{subfigure}
    \begin{subfigure}{0.48\linewidth}
        \centering
        \includegraphics[width=0.98\linewidth]{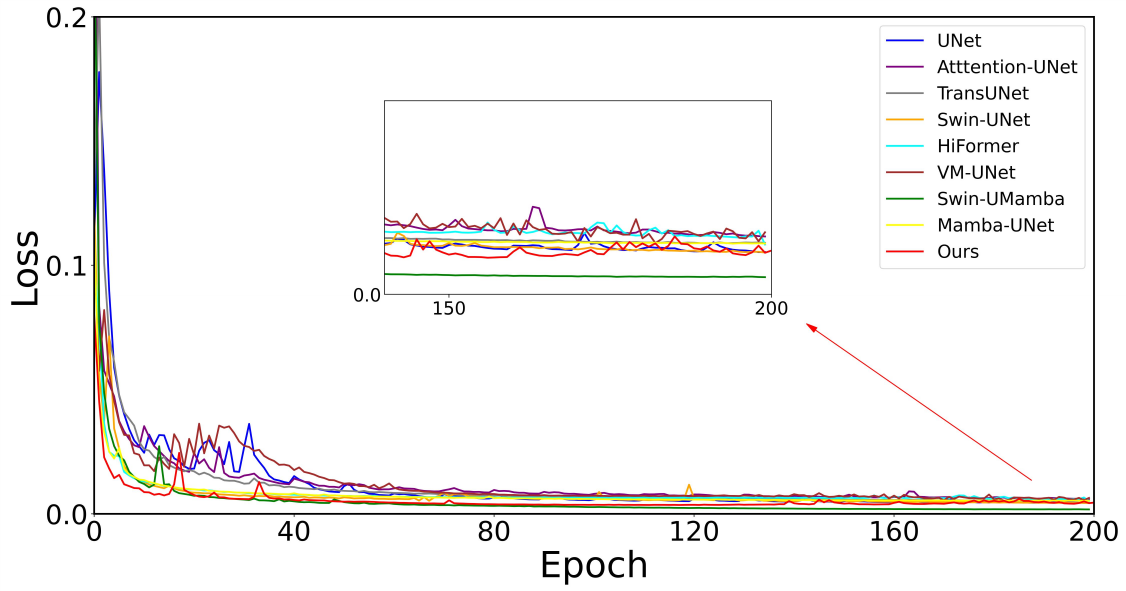}
        \caption{SLIVER}
    \end{subfigure}
    \begin{subfigure}{0.48\linewidth}
        \centering
        \includegraphics[width=0.98\linewidth]{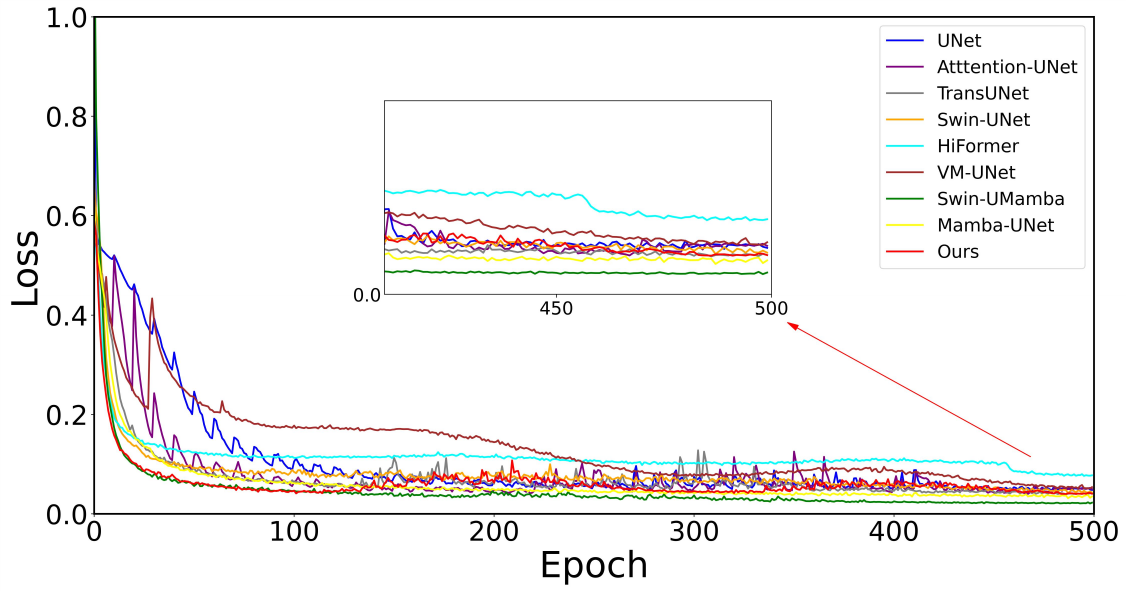}
        \caption{BTCV}
    \end{subfigure}
    \begin{subfigure}{0.48\linewidth}
        \centering
        \includegraphics[width=0.98\linewidth]{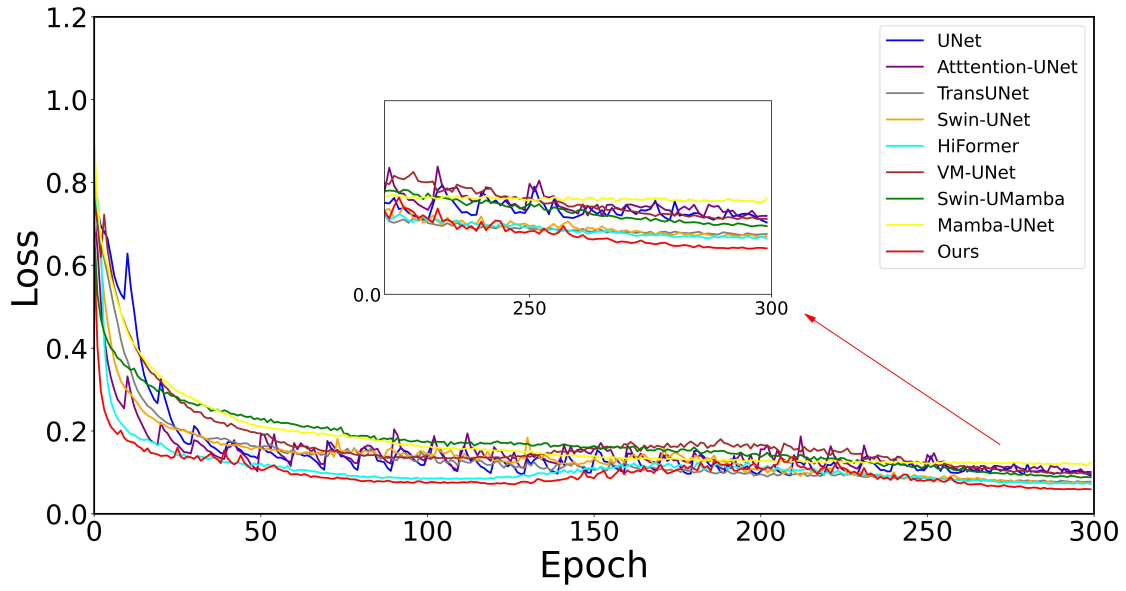}
        \caption{RAOS}
    \end{subfigure}
    \caption{The training loss trends of the proposed method and eight comparative methods on the four datasets.}
\end{figure*}

\subsection{Datasets}
1) \textbf{Synapse Multi-Organ Segmentation Dataset}\cite{b31}: This dataset originates from the MICCAI 2015 Multi-Atlas Abdominal Labeling Challenge and includes abdominal CT scans of 30 patients.  More specifically, Synapse provides annotations for eight abdominal organs and includes 3,779 axial contrast-enhanced clinical CT images.  

2) \textbf{MM-WHS}\cite{b32}: The Multi-Modality Whole Heart Segmentation dataset originates from the MICCAI 2017 Challenge. This dataset encompasses images of the entire heart and its important substructures, providing manual annotations for seven primary cardiac substructures. The dataset includes 20 CT cases and 20 MRI cases.

3) \textbf{CHAOS}\cite{b30}: The Combined Healthy Abdominal Organ Segmentation is one of the classic benchmarks for abdominal medical image segmentation. It provides paired multimodal CT and MRI data. In this dataset, the MRI data includes annotated images for four regions: Liver, Right Kidney, Left Kidney, and Spleen, while only the liver is annotated in the CT data.

4) \textbf{BTCV}\cite{b31}: The Multi-Atlas Labeling Beyond the Cranial Vault dataset originates from a workshop held at MICCAI 2015 and serves as an essential benchmark for abdominal organ segmentation tasks. The in-plane resolution of the images in the dataset varies from 0.54\(\times\)0.54 mm² to 0.98\(\times\)0.98 mm², with slice thickness ranging from 2.5 mm to 5 mm. The dataset includes 3D CT data from 30 patients and provides manual annotations for 13 abdominal organs.

5) \textbf{RAOS}\cite{b34}: Rethinking Abdominal Organ Segmentation serves as a robust evaluation benchmark for challenging cases. This dataset comprises 413 actual clinical CT scans and 413\(\times\)9 MRI scans, with the 19 organs manually annotated by a senior oncologist. Due to time and computational resource constraints in the experimental phase, we select CT data of 40 patients to form the dataset.

6) \textbf{SLIVER}\cite{b33}: The Segmentation of The Liver Competition 2007 is a classic dataset used for liver segmentation in contrast-enhanced CT images, initially used in the MICCAI 2007 Challenge. In CT imaging, liver segmentation serves as the foundation for computer-assisted surgical planning, such as tumour resection, liver transplantation, or minimally invasive procedures. The dataset includes liver CT data from 20 patients.

7) \textbf{LungCancer}\cite{b37}: This dataset consists of chest CT images specifically designed to display and assess the characteristics of pulmonary nod lesions. The dataset highlights variations in lung nodules' morphology, density, size, and growth location. The dataset is divided into two subsets: The first subset contains 300 images of lung nodules that are surgically resected and pathologically diagnosed as malignant tumours. The second subset includes 260 images of lung nodules that are surgically resected and pathologically diagnosed as benign tumours.

\subsection{Evaluation Metrics}
Due to the varying modalities and segmentation regions across the seven datasets, the evaluation metrics employed differ for each dataset. For the CHAOS, Synapse, MM-WHS, BTCV, and RAOS datasets, we primarily utilize the average Dice Similarity Coefficient (DSC) and the average Hausdorff Distance at the 95\(th\) percentile (HD95) as performance evaluation metrics. For the single-label datasets SLIVER and LungCancer, we comprehensively assess model performance using the average DSC, average HD95, average Recall, and average Precision.

\begin{figure*}[h]
    \centering
    \includegraphics[width=0.98\textwidth]{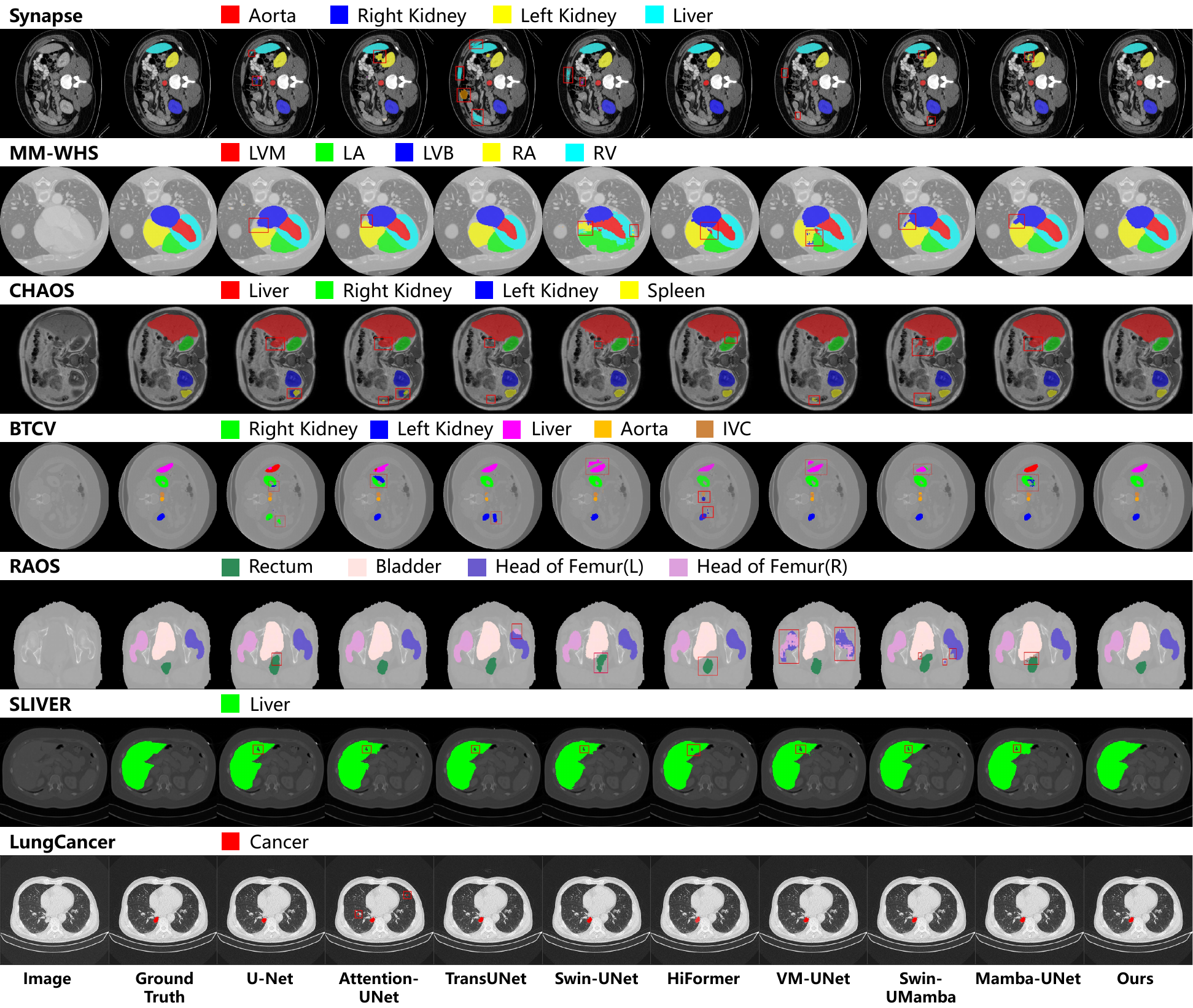}
    \caption{The visualization results of LamFormer and eight comparison methods for image segmentation on seven datasets, as well as the original image and GroundTruth. Incorrect segmentation areas are marked with red boxes.}
\end{figure*}

\subsection{Implementation Detail}
We implement the designed network using PyTorch on an NVIDIA GeForce RTX 3090 with 24GB of memory. To avoid overfitting, a series of data augmentation operations are applied to the dataset before inputting the data into the network. LamFormer is trained from scratch using the AdamW optimizer, with an initial learning rate of 0.001. The loss function throughout the training process was a combination of Cross-Entropy and Dice loss, weighted according to specified parameters. The calculation method is as follows:
\begin{equation}
\begin{aligned}
L(Y,P) &= 1 - \sum_{i=1}^I (\lambda \frac{2 \ast \sum_{n=1}^N Y_{n,i} \cdot P_{n,i}}
{\sum_{n=1}^N Y_{n,i}^2 + \sum_{n=1}^N P_{n,i}^2} \\
       &+ \sum_{n=1}^N Y_{n,i}log P_{n,i}),
\end{aligned}
\end{equation}
where \(I\) represents the number of classes, \(N\) denotes the total number of voxels, and  \(P(n, i)\) and \(Y(n, i)\) corresponds to the predicted output and the ground truth for class \(i\) at voxel \(v\). We conducted various experiments and presented the average results. Statistical analysis indicates that our method significantly outperforms comparative methods.

\subsection{Comparison With Other Methods}
We employ eight state-of-the-art methods to validate the performance of LamFormer: U-Net (2015), Attention-UNet (2018), TransUNet (2021), Swin-UNet (2022), HiFormer (2023), VM-UNet (2024), Swin-UMamba (2024) and Mamba-UNet (2024). As shown in Figure 5, we plot the loss curves of all methods during the training process on four datasets.

\begin{table*}[h]
\centering
\captionsetup{justification=centering, labelformat=empty}
\caption*{\textcolor[RGB]{0,150,220}{TABLE \uppercase\expandafter{\romannumeral1}}\\COMPARISONS WITH STATE-OF-THE-ART MODELS ON THE SYNAPSE MULTI-ORGAN SEGMENTATION DATASET. THE PERFORMANCES ON SEGMENTING THE EIGHT ORGANS ARE ALL REPORTED.}
\begin{tabular}{ccccccccccc}
\toprule
\multirow{2}*{Method} & DSC\textcolor{red}{↑} & HD95\textcolor{red}{↓} & Aorta & Gallbladder & Right kidney & Left kidney & Liver & Pancreas & Spleen & Stomach \\
& (\%, mean) & (mm, mean) & \multicolumn{8}{c}{DSC\textcolor{red}{↑}} \\
\midrule
U-Net & 77.47 & 29.955 & 87.49 & 69.25 & 82.78 & 73.74 & 92.87 & 57.02 & 87.52 & 69.02 \\
Attention-UNet & 76.67 & 33.894 & 86.64 & 66.67 & 80.91 & 74.58 & 91.91 & 56.81 & 85.14 & 70.63 \\
TransUNet & 79.66 & 29.967 & 86.71 & 66.62 & 85.12 & 80.85 & 93.62 & 60.96 & 89.26 & 74.11 \\
Swin-UNet & 78.08 & 23.949 & 82.50 & 66.47 & 83.16 & 80.40 & 92.94 & 59.32 & 87.47 & 72.36 \\
HiFormer & 78.97 & 26.868 & 84.96 & 65.80 & 81.85 & 79.34 & 94.24 & 56.56 & 88.99 & \textbf{80.06} \\
VM-UNet & 79.93 & 23.484 & 84.65 & 67.21 & 82.19 & 73.91 & 92.93 & 54.72 & 88.25 & 74.60 \\
Swin-UMamba & 80.19 & 18.850 & 86.73 & 62.59 & 84.20 & 79.50 & 93.90 & 49.93 & 87.20 & 69.83 \\
Mamba-UNet & 79.68 & 25.710 & 86.81 & 67.13 & 85.03 & \textbf{81.84} & 93.77 & 55.85 & 87.03 & 77.06 \\
Ours & \textbf{81.22} & \textbf{16.998} & \textbf{87.57} & \textbf{69.33} & \textbf{85.26} & 81.45 & \textbf{94.16} & \textbf{62.26} & \textbf{90.52} & 79.20 \\
\bottomrule
\end{tabular}
\end{table*}

\begin{table*}[h]
\centering
\captionsetup{justification=centering, labelformat=empty}
\caption*{\textcolor[RGB]{0,150,220}{TABLE \uppercase\expandafter{\romannumeral2}}\\COMPARISONS WITH STATE-OF-THE-ART MODELS ON THE MM-WHS DATASET.}
\begin{tabular}{cccccccccc}
\toprule
\multirow{2}*{Method} & DSC\textcolor{red}{↑} & HD95\textcolor{red}{↓} & LVM & LA & LVB & RA & RV & Ao & PA \\
 & (\%, mean) & (mm, mean) & \multicolumn{7}{c}{DSC\textcolor{red}{↑}} \\
\midrule
U-Net & 76.83 & 34.980 & 82.04 & 76.30 & 78.29 & 80.60 & 75.97 & 72.77 & 71.83 \\
Attention-UNet & 78.49 & 22.306 & 83.07 & 77.23 & 80.84 & 83.33 & 78.59 & 73.67 & 72.71 \\
TransUNet & 79.08 & 20.469 & 82.74 & 77.84 & 83.69 & 84.82 & 77.36 & 76.37 & 70.76 \\
Swin-UNet & 64.27 & 34.938 & 68.70 & 64.69 & 66.63 & 72.56 & 65.61 & 61.95 & 49.76 \\
HiFormer & 79.56 & 23.016 & 84.50 & 78.09 & 84.10 & 84.15 & 78.83 & 74.92 & 72.34 \\
VM-UNet & 77.96 & 24.911 & 83.64 & 72.05 & 82.02 & 80.39 & 81.71 & 77.45 & 68.45 \\
Swin-UMamba & 79.35 & 29.098 & 83.10 & 78.29 & 85.82 & 82.75 & 79.06 & 76.02 & 70.41 \\
Mamba-UNet & 80.35 & 20.377 & 85.92 & 76.52 & 85.38 & 85.35 & 80.23 & \textbf{79.98} & 69.10 \\
Ours & \textbf{83.02} & \textbf{14.569} & \textbf{88.75} & \textbf{81.32} & \textbf{88.38} & \textbf{86.98} & \textbf{84.63} & 77.79 & \textbf{73.68} \\
\bottomrule
\end{tabular}
\end{table*}

\begin{table*}[h]
\centering
\captionsetup{justification=centering, labelformat=empty}
\caption{\textcolor[RGB]{0,150,220}{TABLE \uppercase\expandafter{\romannumeral3}}\\COMPARISONS WITH STATE-OF-THE-ART MODELS ON THE CHAOS DATASET.}
\begin{tabular}{ccccccc}
\toprule
\multirow{2}*{Method} & DSC\textcolor{red}{↑} & HD95\textcolor{red}{↓} & Liver & Right kidney & Left kidney & Spleen \\
 & (\%, mean) & (mm, mean) & \multicolumn{4}{c}{DSC\textcolor{red}{↑}} \\
\midrule
U-Net & 90.50 & 6.997 & 94.44 & 92.47 & 89.99 & 85.10 \\
Attention-UNet & 90.26 & 8.687 & 93.29 & 89.31 & 90.34 & 88.10 \\
TransUNet & 90.62 & 5.224 & 94.10 & 91.38 & 89.76 & 87.24 \\
Swin-UNet & 90.42 & 3.239 & 93.18 & 90.44 & 90.86 & 87.22 \\
HiFormer & 90.90 & \textbf{2.513} & 93.65 & 91.90 & 89.26 & 88.78 \\
VM-UNet & 90.56 & 3.182 & 92.98 & 91.99 & 89.54 & 87.74 \\
Swin-UMamba & 91.58 & 3.787 & 94.41 & 93.53 & 92.15 & 86.23 \\
Mamba-UNet & 91.30 & 3.936 & 93.85 & 90.94 & 91.30 & \textbf{89.10} \\
Ours & \textbf{92.09} & 3.492 & \textbf{94.65} & \textbf{93.67} & \textbf{92.69} & 87.54 \\
\bottomrule
\end{tabular}
\end{table*}

1) \textbf{Experiments on the Synapse dataset}: Table \uppercase\expandafter{\romannumeral1} presents the results for the Synapse dataset, where Swin-UMamba outperforms other existing methods. However, LamFormer performs better in DSC and HD95 than the contrastive models. In particular, the segmentation performance for the aorta, gallbladder, right kidney, liver, pancreas, and spleen is the best.
\par The first row of Fig. 6 visually displays the segmentation outcomes. Incorrectly segmented regions are highlighted with red boxes. Existing methods exhibit ambiguous boundaries for the liver and left kidney, leading to confusion with the pancreas, spleen, stomach, and liver. In contrast, our method more accurately delineates the boundaries among these organs.

2) \textbf{Experiments on the MM-WHS dataset}: Table \uppercase\expandafter{\romannumeral2} shows the results of the MM-WHS dataset. U-Net achieves a DSC of 76.83\%, while HiFormer achieves a DSC of 79.56\%. Furthermore, Mamba-UNet achieves a DSC of 80.35\%, outperforming existing methods. LamFormer achieves a DSC of 83.02\%, surpassing Mamba-UNet.
\par The second row of Fig. 6 shows the segmentation results. UNet, Attention-UNet, Swin-UMamba, and Mamba-UNet make segmentation errors on LVB and RA, while Swin-UNet, HiFormer, and VM-UNet make segmentation errors on LA, RA, and RV. Existing methods exhibit ambiguous LA, LVB, RA, and RV boundaries. LamFormer more accurately segments all organs and delineates the limits in the examples.

\begin{table*}[h]
\centering
\captionsetup{justification=centering, labelformat=empty}
\caption{\textcolor[RGB]{0,150,220}{TABLE \uppercase\expandafter{\romannumeral4}}\\COMPARISONS WITH STATE-OF-THE-ART MODELS ON THE BTCV DATASET.}
\setlength{\tabcolsep}{3pt}{
\resizebox{\textwidth}{!}{
\begin{tabular}{cccccccccccccccc}
\toprule
\multirow{2}*{Method} & DSC\textcolor{red}{↑} & HD95\textcolor{red}{↓} & Spleen & Kidney(R) & Kidney(L) & Gallbladder & Esophagus & Liver & Stomach & Aorta & IVC & HPV\&SV & Pancreas & Adrenal(L) & Adrenal(R) \\
 & (\%, mean) & (mm, mean) & \multicolumn{13}{c}{DSC\textcolor{red}{↑}} \\
\midrule
U-Net & 67.02 & 21.095 & 82.00 & 67.14 & 75.80 & 56.26 & 64.83 & 92.01 & 74.63 & 84.77 & 68.23 & 56.50 & 52.69 & 45.30 & 51.06\\
Attention-UNet & 72.90 & 28.998 & 83.50 & 70.74 & 83.41 & \textbf{63.11} & 73.23 & 91.94 & 78.27 & 85.02 & 73.74 & \textbf{69.95} & \textbf{66.53} & 54.74 & 57.46 \\
TransUNet & 69.22 & 27.824 & 84.80 & \textbf{83.87} & 83.97 & 56.15 & 70.47 & 92.65 & 72.09 & 83.71 & 67.05	& 60.39	& 50.94	& 41.70	& 52.04 \\
Swin-UNet & 68.49 & 20.096 & 89.54 & 75.26 & 82.26 & 61.91 & 62.82 & 93.55 & 79.30 & 84.25 & 63.16 & 57.96 & 54.94 & 42.68 & 42.81 \\
HiFormer & 65.29 & 16.917 & 86.50 & 68.90 & 77.64 & 62.12 & 65.40 & 93.95 & \textbf{82.31} & 80.89 & 67.85 & 63.01 & 57.37 & 46.57 & 42.84 \\
VM-UNet & 72.70 & 19.866 & 87.69 & 76.85 & 75.26 & 61.60 & 64.75 & 92.90 & 76.99 & 84.77 & 67.68 & 58.80 & 60.05 & 47.20 & 42.59 \\
Swin-UMamba & 73.08 & 19.366 & 88.48 & 80.83 & 80.99 & 54.99 & 65.90 & 93.72 & 69.31 & 81.64 & 66.94 & 56.56 & 50.33 & 44.52 & 53.25 \\
Mamba-UNet & 72.58 & 17.619 & 89.31 & 78.04 & 74.64 & 55.12 & 61.59 & 92.97 & 80.82 & 82.17 & 66.10 & 52.22 & 56.18 & 49.35 & 48.86 \\
Ours & \textbf{74.21} & \textbf{16.002} & \textbf{90.57} & 78.02 & \textbf{87.64} & 55.56 & \textbf{74.99} & \textbf{94.52} & 78.29 & \textbf{87.99} & \textbf{73.75} & 65.20 & 63.67 & \textbf{56.16} & \textbf{58.33} \\
\bottomrule
\end{tabular}
}}
\end{table*}

\begin{table*}[h]
\centering
\captionsetup{justification=centering, labelformat=empty}
\caption*{\textcolor[RGB]{0,150,220}{TABLE \uppercase\expandafter{\romannumeral5}}\\COMPARISONS WITH STATE-OF-THE-ART MODELS ON THE RAOS DATASET.}
\resizebox{\textwidth}{!}{
\begin{tabular}{cccccccccccc}
\toprule
\multirow{2}*{Method} & DSC\textcolor{red}{↑} & HD95\textcolor{red}{↓} & Liver & Spleen & Kidney(L) & Kidney(R) & Stomach & Gallbladder & Esophagus & Pancreas & Duodenum\\
 & (\%, mean) & (mm, mean) & \multicolumn{9}{c}{DSC\textcolor{red}{↑}} \\
\midrule
U-Net & 60.80 & 28.539 & 91.31 & 86.46 & 67.30 & 69.93 & 76.99 & \textbf{58.32} & 59.20 & 51.28 & 41.30 \\
Attention-UNet & 60.63 & 20.685 & 91.22 & 79.33 & 66.89 & 62.60 & 78.24 & 56.34 & 59.16 & 54.28 & 44.52 \\
TransUNet & 63.39 & 17.016 & 91.90 & 91.14 & 81.47 & 82.40 & 77.19 & 55.52 & 63.74 & \textbf{67.23} & 46.64 \\
Swin-UNet & 64.09 & 26.417 & 92.73 & 88.57 & 77.96 & 80.77 & 77.10 & 44.59 & 57.71 & 54.95 & 33.98 \\
HiFormer & 66.21 & 17.634 & 92.58 & 84.08 & 79.54 & 79.45 & 76.37 & 56.68 & 62.78 & 55.15 & 40.42 \\
VM-UNet & 64.75 & 18.609 & 92.78 & 90.87 & 74.63 & 65.20 & 77.93 & 55.28 & 61.69 & 59.75 & 46.51 \\
Swin-UMamba & 68.78 & 19.822 & 90.39 & 87.85 & 81.19 & \textbf{86.15} & 65.05 & 34.78 & 64.12 & 60.13 & 40.63 \\
Mamba-UNet & 66.35 & 26.163 & 92.00 & 89.15 & 69.72 & 71.38 & 77.43 & 44.19 & 55.05 & 56.59 & 39.57 \\
Ours & \textbf{70.20} & \textbf{16.626} & \textbf{92.93} & \textbf{91.41} & \textbf{82.61} & 67.58 & \textbf{78.71} & 51.84 & \textbf{64.14} & 55.53 & \textbf{47.33}\\
\bottomrule
\end{tabular}
}
\resizebox{\textwidth}{!}{
\begin{tabular}{ccccccccccc}
\\
\toprule
\multirow{2}*{Method} & Colon & Intestine & Adrenal(L) & Adrenal(R) & Rectum & Bladder & HoF(L) & HoF(R) & Prostate & SV \\
 & \multicolumn{10}{c}{DSC\textcolor{red}{↑}} \\
\midrule
U-Net & 69.71 & 73.17 & 0.00 & 0.00 & 76.46 & 91.99 & 88.24 & 88.34 & 65.24 & 0.00 \\
Attention-UNet & 73.14 & 76.03 & 0.00 & 0.00 & 76.21 & 91.71 & 84.18 & 88.25 & 69.86 & 0.00 \\
TransUNet & 72.37 & 77.31 & 0.00 & 32.30 & 74.72 & 89.53 & \textbf{92.39} & \textbf{92.08} & 16.47 & 0.00 \\
Swin-UNet & 71.73 & 74.38 & 41.96 & 34.79 & 72.38 & 91.81 & 84.35 & 81.25 & 37.85 & 18.84 \\
HiFormer & 71.33 & 75.15 & 0.00 & 47.44 & 78.73 & 93.34 & 87.08 & 86.47 & 57.06 & 34.40 \\
VM-UNet & \textbf{73.23} & 74.59 & 57.50 & 46.05 & 76.40 & 92.55 & 66.00 & 66.49 & 52.85 & 0.00 \\
Swin-UMamba & 66.51 & 67.93 & 44.86 & 47.11 & 72.01 & 88.18 & 89.97 & 90.84 & \textbf{74.29} & 54.83 \\
Mamba-UNet & 67.59 & 70.77 & 45.89 & 36.10 & 69.26 & 92.46 & 78.33 & 77.49 & 61.81 & \textbf{65.83} \\
Ours & 71.34 & \textbf{77.71} & \textbf{58.80} & \textbf{48.01} & \textbf{79.15} & \textbf{94.58} & 85.81 & 85.70 & 59.59 & 40.98 \\
\bottomrule
\end{tabular}}
\end{table*}

3) \textbf{Experiments on the CHAOS dataset}: Table \uppercase\expandafter{\romannumeral3} shows the results of the CHAOS dataset. Compared to the best-performing existing method, Swin-UMamba, LamFormer achieves a DSC of 92.09\% and demonstrates superior segmentation performance for the liver, left kidney, and right kidney.
\par The third row of Fig. 6 visually compares our method with other methods. UNet and Attention-UNet make errors in the spleen segmentation, confusing the left kidney and spleen. While existing methods can segment the right kidney, left kidney, and spleen, some mistakenly classify the background as liver. UNet, Attention-UNet, Swin-UMamba, and Mamba-UNet exhibit significant misidentification in the liver region, whereas our method improves the accuracy of liver segmentation.

4) \textbf{Experiments on the BTCV dataset}: Table \uppercase\expandafter{\romannumeral4} presents the results of the BTCV dataset. Our method performs better with DSC and HD95 scores of 74.21\% and 16.002, outperforming existing methods. Optimal metrics are achieved for the spleen, left kidney, esophagus, liver, aorta, left adrenal gland, and right adrenal gland.
\par In Fig. 6, row four illustrates a visual comparison of segmentation outcomes. U-Net, Attention-UNet, Swin-UMamba, and Mamba-UNet exhibit segmentation errors in the right kidney and liver. At the same time, U-Net, TransUNet, Swin-UNet, HiFormer, and VM-UNet incorrectly delineate the background.

5) \textbf{Experiments on the RAOS dataset}: Since this dataset contains 19 organ categories, the experimental results are divided into two parts in Table \uppercase\expandafter{\romannumeral5}. Swin-UMamba is the best-performing existing method, achieving a DSC of 68.78\%. Our method surpasses Swin-UMamba in DSC and HD95, significantly outperforming other methods. Due to the more segmentation categories, LamFormer cannot achieve optimal performance among all 19 organ classes. Besides, the complexity and diversity of segmented organs make it challenging to identify organs with low contrast and small size. TransUNet, HiFormer, and Swin-UMamba fail to segment the left and right adrenal glands.
\par As shown in Fig. 6, U-Net, Swin-UNet, and Mamba-UNet exhibit errors in rectum segmentation. TransUNet and VM-UNet make confusion errors when segmenting the left and right femoral heads. Our method demonstrates superior performance. However, due to the numerous categories in the dataset, there remains considerable potential for promoting segmentation performance.

6) \textbf{Experiments on the SLIVER dataset}: Since previous experiments primarily focus on multi-organ/multi-region segmentation, we chose a single-organ dataset for segmentation in this study. The objective is to validate our model's effectiveness for segmenting individual organs/regions. Table \uppercase\expandafter{\romannumeral6} shows the experimental results. LamFormer significantly outperforms compared methods on DSC, HD95, Recall, and Precision metrics.
\par The sixth row of Fig. 6 provides a visual comparison between LamFormer and compared methods. Existing methods can roughly segment the liver. However, they fall short in terms of detail. Notably, LamFormer produces more accurate segmentation results.

7) \textbf{Experiments on the LungCancer dataset}: As mentioned earlier, the experiments conducted on the datasets focus on organ segmentation. Therefore, we select a lung lesion
dataset to validate the generalization ability of LamFormer for small target segmentation. According to Table \uppercase\expandafter{\romannumeral7}, Swin-UMamba achieves the best DSC, Recall, and Precision metrics among existing methods, with values of 70.75\%, 71.65\%, and 74.29\%. Swin-UNet achieves the best HD95 metric among existing methods, measuring 5.051 mm. Subsequently, our method surpasses all comparative methods in DSC, HD95, Recall, and Precision, notably achieving a DSC of 72.04\%.
\par The seventh row of Fig. 6 visually compares LamFormer with eight comparative methods.Although comparative methods can capture the rough shape of lung cancer, they display limited smoothness in handling edges. Overall, LamFormer exhibits superior segmentation performance.

\begin{table*}[!t]
\begin{minipage}{0.5\linewidth}
\captionsetup{justification=centering, labelformat=empty}
\caption*{\textcolor[RGB]{0,150,220}{TABLE \uppercase\expandafter{\romannumeral6}}\\COMPARISONS WITH STATE-OF-THE-ART MODELS \quad ON THE SLIVER DATASET.} 
\begin{tabular}{ccccc}
\toprule
\multirow{2}*{Method} & DSC\textcolor{red}{↑} & HD95\textcolor{red}{↓} & Recall\textcolor{red}{↑} & Precision\textcolor{red}{↑} \\
 & (\%, mean) & (mm, mean) & (\%, mean) & (\%, mean) \\
\midrule
U-Net & 93.98 & 16.612 & 93.05 & 95.08 \\
Attention-UNet & 93.50 & 23.168 & 93.75 & 93.45 \\
TransUNet & 93.96 & 19.027 & 93.64 & 94.46 \\
Swin-UNet & 94.88 & 17.522 & 94.64 & 95.21 \\
HiFormer & 95.77 & 13.524 & 95.82 & 95.75 \\
VM-UNet & 95.53 & 15.734 & 94.85 & 96.35 \\
Swin-UMamba & 95.55 & 7.645 & 96.01 & 96.15 \\
Mamba-UNet & 95.60 & 9.088 & 95.95 & 96.33 \\
Ours & \textbf{96.97} & \textbf{6.645} & \textbf{96.38} & \textbf{97.59} \\
\bottomrule
\end{tabular}
\end{minipage}%
\begin{minipage}{0.5\linewidth}
\captionsetup{justification=centering, labelformat=empty}
\caption*{\textcolor[RGB]{0,150,220}{TABLE \uppercase\expandafter{\romannumeral7}}\\COMPARISONS WITH STATE-OF-THE-ART MODELS \quad ON THE LungCancer DATASET.}
\begin{tabular}{ccccc}
\toprule
\multirow{2}*{Method} & DSC\textcolor{red}{↑} & HD95\textcolor{red}{↓} & Recall\textcolor{red}{↑} & Precision\textcolor{red}{↑} \\
 & (\%, mean) & (mm, mean) & (\%, mean) & (\%, mean) \\
\midrule
U-Net & 68.73 & 21.303 & 69.19 & 71.15 \\
Attention-UNet & 65.73 & 19.262	& 65.68	& 70.26 \\
TransUNet & 67.56 & 10.162 & 67.30 & 70.25 \\
Swin-UNet & 65.26 & 5.051 & 64.62 & 68.72 \\
HiFormer & 67.30 & 8.608 & 69.90 & 69.65 \\
VM-UNet & 69.68 & 18.149 & 70.76 & 73.38 \\
Swin-UMamba & 70.75 & 10.446 & 71.65 & 74.29 \\
Mamba-UNet & 70.19 & 8.081 & 70.09 & 73.10 \\
Ours & \textbf{72.04} & \textbf{4.299} & \textbf{71.96} & \textbf{75.93} \\
\bottomrule
\end{tabular}
\end{minipage}%
\end{table*}

\begin{table*}[h]
\centering
\captionsetup{justification=centering, labelformat=empty}
\caption*{\textcolor[RGB]{0,150,220}{TABLE \uppercase\expandafter{\romannumeral8}}\\ABLATION EXPERIMENTS ON THE MM-WHS DATASET.}
\begin{tabular}{cccccccccc}
\toprule
\multirow{2}*{Method} & DSC\textcolor{red}{↑} & HD95\textcolor{red}{↓} & LVM & LA & LVB & RA & RV & Ao & PA \\
 & (\%, mean) & (mm, mean) & \multicolumn{7}{c}{DSC\textcolor{red}{↑}} \\
\midrule
LAM & 81.64 & 19.234 & 87.25 & 79.99 & 86.53 & 86.56 & 84.31 & 77.96 & 68.88 \\
PHFA & 81.86 & 22.939 & 88.38 & 80.00 & 87.61 & 85.95 & 84.11 & 72.78 & 72.84 \\
RT & 81.55 & 18.355 & 88.12 & 80.85 & 85.43 & 85.62 & 84.56 & 77.26 & 69.01 \\
LAM+PHFA & 82.53 & 15.821 & 87.12 & 81.12 & 86.41 & 85.07 & 84.04 & \textbf{80.40} & 73.56 \\
LAM+RT & 82.45 & 15.887 & 87.50 & 81.29 & 88.27 & 86.36 & 83.61 & 77.67 & 72.47 \\
PHFA+RT & 82.27 & 16.374 & 88.31 & 81.01 & 86.82 & 85.99 & \textbf{84.77} & 78.44 & 70.58 \\
LAM+PHFA+RT & \textbf{83.02} & \textbf{14.569} & \textbf{88.75} & \textbf{81.32} & \textbf{88.38} & \textbf{86.98} & 84.63 & 77.79 & \textbf{73.68} \\
\bottomrule
\end{tabular}
\end{table*}

\begin{table}[h]
\centering
\captionsetup{justification=centering, labelformat=empty}
\caption*{\textcolor[RGB]{0,150,220}{TABLE \uppercase\expandafter{\romannumeral9}}\\ABLATION EXPERIMENTS ON THE SLIVER \quad \quad DATASET.}
\begin{tabular}{ccccc}
\toprule
\multirow{2}*{Method} & DSC\textcolor{red}{↑} & HD95\textcolor{red}{↓} & Recall\textcolor{red}{↑} & Precision\textcolor{red}{↑} \\
 & (\%, mean) & (mm, mean) & (\%, mean) & (\%, mean) \\
\midrule
LAM & 95.15 & 13.774 & 95.40 & 94.97 \\
PHFA & 95.38 & 11.281 & 94.65 & 96.21 \\
RT & 94.97 & 10.997 & 94.10 & 95.92 \\
LAM+PHFA & 96.44 & 7.065 & 96.10 & 96.83 \\
LAM+RT & 96.59 & 7.310 & 96.13 & 97.08 \\
RT+PHFA  & 95.42 & 9.349 & 95.52 & 95.39 \\
LAM+PHFA+RT & \textbf{96.97} & \textbf{6.645} & \textbf{96.38} & \textbf{97.59} \\
\bottomrule
\end{tabular}
\end{table}

\begin{table}[h]
\centering
\captionsetup{justification=centering, labelformat=empty}
\caption*{\textcolor[RGB]{0,150,220}{TABLE \uppercase\expandafter{\romannumeral10}}\\ABLATION EXPERIMENTS ON THE LungCancer DATASET.}
\begin{tabular}{ccccc}
\toprule
\multirow{2}*{Method} & DSC\textcolor{red}{↑} & HD95\textcolor{red}{↓} & Recall\textcolor{red}{↑} & Precision\textcolor{red}{↑} \\
 & (\%, mean) & (mm, mean) & (\%, mean) & (\%, mean) \\
\midrule
LAM & 69.42 & 11.763 & 68.68 & 75.50 \\
PHFA & 68.82 & 16.367 & 68.07 & 72.35 \\
RT & 68.24 & 10.585 & 67.28 & 71.99 \\
LAM+PHFA & 71.69 & 6.059 & 71.41 & 74.11 \\
LAM+RT & 70.78 & 13.376 & 71.15 & 74.34 \\
PHFA+RT & 70.62 & 5.491 & 69.97 & 74.65 \\
LAM+PHFA+RT & \textbf{72.04} & \textbf{4.299} & \textbf{71.96} & \textbf{75.93} \\
\bottomrule
\end{tabular}
\end{table}

\subsection{Ablation Studies}
Our method mainly introduces three components: the Linear Attention Mamba (LAM), the Parallel Hierarchical Feature Aggregation (PHFA), and the Reduced Transformer (RT). We conducted ablation experiments on the MM-WHS, SLIVER, and LungCancer datasets. As shown in Tables VIII, IX, and X, ``LAM+PHFA+RT" generally surpasses all other combinations on all datasets, demonstrating the effectiveness of this particular combination. First, we conduct experiments on three datasets using each of the LAM, PHFA, and RT modules individually. Next,  we adopt a two-by-two combination approach of the three modules, resulting in further improved segmentation results on all three datasets. Finally, our segmentation performance reaches its best when we use all three modules: LAM, PHFA, and RT.
\par Overall, our approach achieves superior performance in all segmentation results on the SLIVER and LungCancer datasets. Additionally, on the MM-WHS dataset, our method demonstrates improved segmentation results in five out of seven organs.

\begin{figure}[h]
    \centering
    \includegraphics[width=0.47\textwidth]{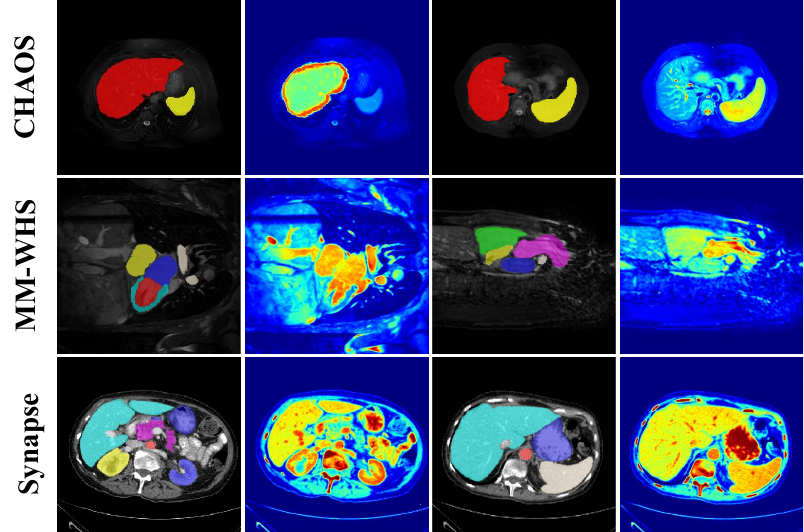}
    \caption{The visualization images of attention-weight heatmaps on multiple datasets.}
\end{figure}

\subsection{Discussions}
1) \textbf{Discussion on the image characteristics}: We use attention-weight heatmap visualization to visually illustrate the performance differences of the designed model across various datasets. As shown in Fig. 7, the designed model accurately captures the regions of interest in the Synapse dataset, resulting in optimal segmentation performance. However, the dense distribution of the eight segmentation labels and minimal differences in image features adversely affect the segmentation performance. In the CHAOS dataset, the differences in segmentation features are more pronounced, and the segmented regions are more dispersed, leading to the model's superior segmentation performance. On the MM-WHS dataset, despite the dense segmentation regions, the high quality of the dataset enables the designed model to achieve significant segmentation performance.
\par By visualizing attention weight heatmaps, we can observe that LamFormer focuses on segmentation target, highlighting the importance of target organ.

\begin{figure}[h]
    \centering
    \includegraphics[width=0.47\textwidth]{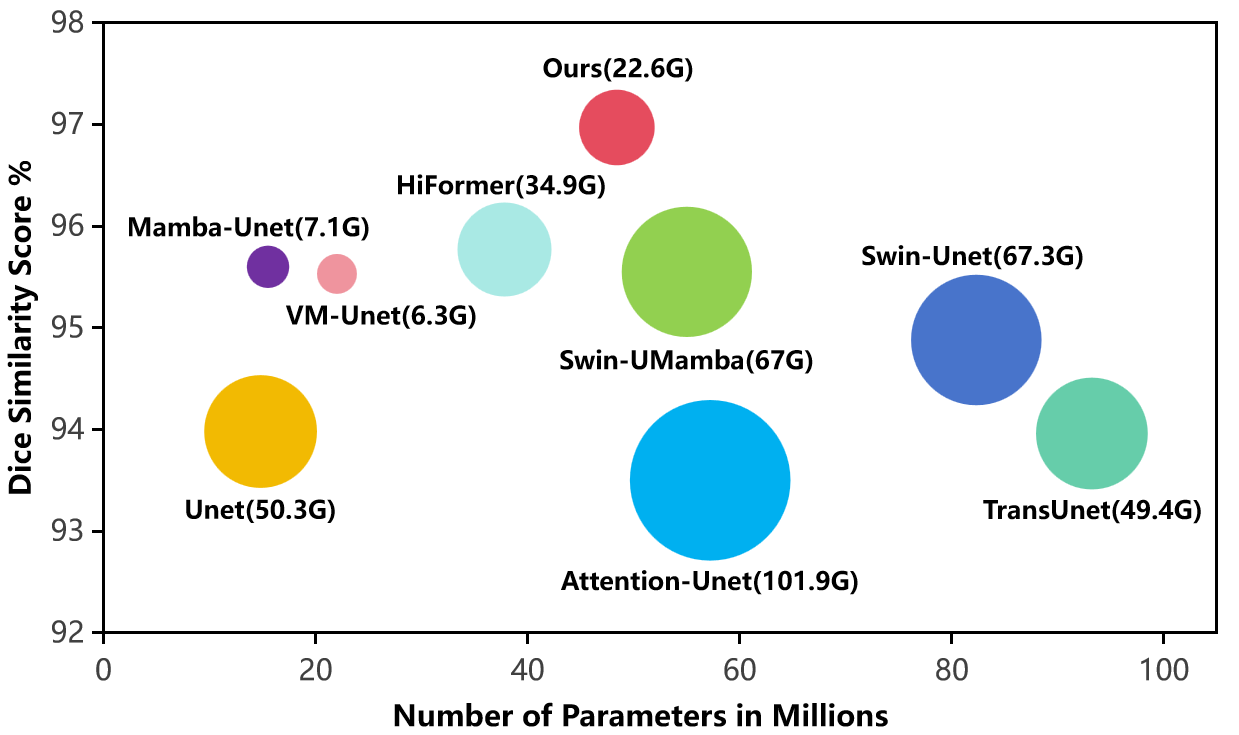}
    \caption{Accuracy (Dice) vs. model parameters(Params) and Floating Point Operations(FLOPs) comparison on the SLIVER dataset.}
\end{figure}

2) \textbf{Discussion on the model complexity}: In this study, we investigate the model parameters (Params) and Floating Point Operations (FLOPs). The Params and FLOPs are computed with input resolution of 224 × 224. From Fig. 8, it can be seen that the FLOPs of LamFormer are significantly lower than those of most of the compared methods, but are slightly higher than those of VM-UNet and Mamba-UNet. Moreover, LamFormer exhibits superior performance while balancing the number of parameters. The above experiments demonstrate that our model can be utilized in scenarios with limited resources and exhibits superior performance.

\begin{figure}[h]
    \centering
    \includegraphics[width=0.45\textwidth]{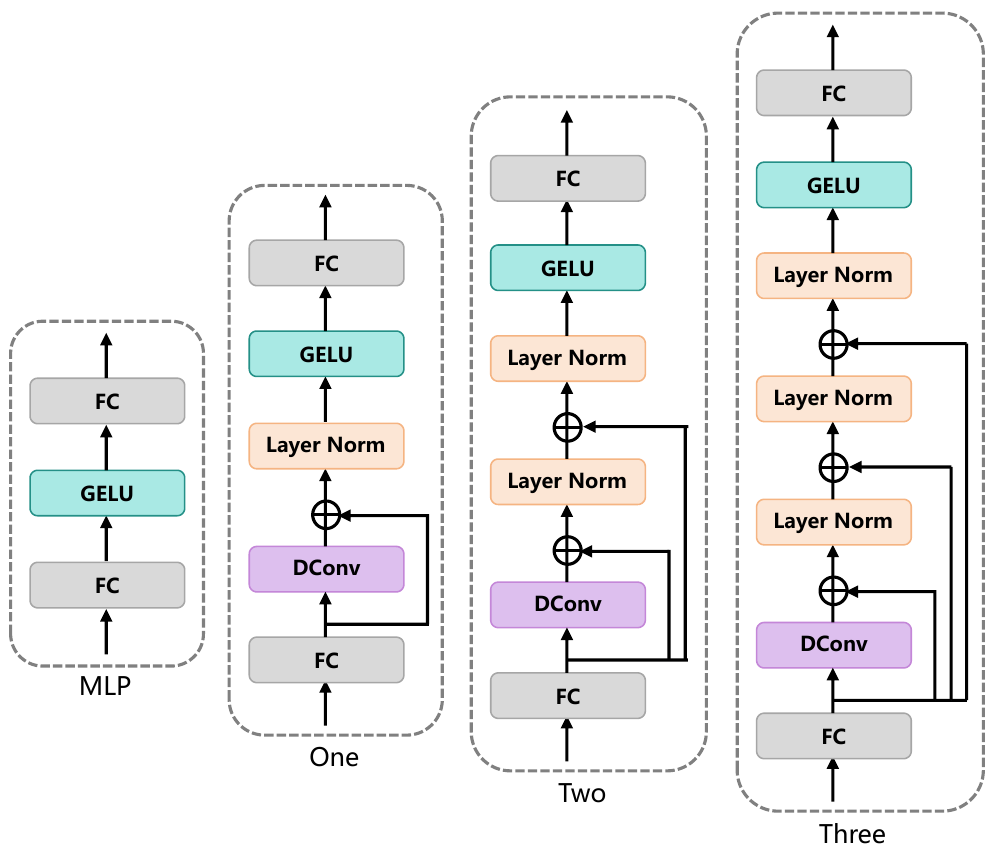}
    \caption{The differences between four forward feedback networks.}
    \label{fig:enter-label}
\end{figure}

3) \textbf{Discussion on the configurations of Forward Residual Network}: We conduct experiments on four datasets using MLP, Forward Residual Network (FRN), and variants of FRN to verify the performance of FRN. As shown in Figure 9, the structures of MLP, FRN and variants of FRN are shown respectively. ``One" denotes the use of a single layer normalization and a single residual connection in the FRN. ``Two" represents the use of two layer normalizations and two residual connections. Table \uppercase\expandafter{\romannumeral11} shows that FRN and its variants are generally better than MLP, and FRN achieves the best performance. This demonstrates that the FRN employing three layer normalizations and three residual connections exhibits superior robustness and generalization ability across different datasets.

\begin{table}[h]
\centering
\captionsetup{justification=centering, labelformat=empty}
\caption*{\textcolor[RGB]{0,150,220}{TABLE \uppercase\expandafter{\romannumeral11}}\\EXPERIMENTS ON THE FORWARD RESIDUAL NETWORK(FRN).}
\begin{tabular}{ccccc}
\toprule
\multirow{2}*{Method} & CHAOS & Synapse & SLIVER & LungCancer \\ &\multicolumn{4}{c}{DSC\textcolor{red}{↑}(\%, mean)} \\
\midrule
MLP & 91.30 & 79.85 & 96.07 & 66.99\\
One & 91.57 & 80.63 & 96.68 & 71.69\\
Two & 91.89 & 80.82 & 96.52 & 71.86\\
Three(Ours) & \textbf{92.09} & \textbf{81.22} & \textbf{96.97} & \textbf{72.04}\\
\bottomrule
\end{tabular}
\end{table}

4) \textbf{Discussion on the inference time and GPU memory utilization}: As shown in Table \uppercase\expandafter{\romannumeral12}, ``SA" represents Self-Attention in Transformer, ``RSA" represents Reduced Self-Attention in Reduced Transformer, and ``LA" represents linear attention in Linear Attention Mamba. We calculated the inference time, GPU memory utilization, and segmentation performance dataset for each of them as well as their combinations. ``LA" achieves the lowest inference time and GPU memory utilization, while LamFormer uses the combination of ``LA+RSA" to achieve the best segmentation performance at the cost of some inference time and GPU memory utilization.
\par Overall, our method optimizes the model architecture to effectively balance performance, complexity, and inference time. This demonstrates its immense potential for clinical practice.

\begin{table}[h]
\centering
\captionsetup{justification=centering, labelformat=empty}
\caption*{\textcolor[RGB]{0,150,220}{TABLE \uppercase\expandafter{\romannumeral12}}\\THE DISCUSSION ON THE INFERENCE TIME AND GPU MEMORY UTILIZATION.}
\begin{tabular}{cccc}
\toprule
\multirow{2}*{Method} & Inference & GPU Mem Util & CHAOS \\ & (ms\textcolor{red}{↓}) & (GB\textcolor{red}{↓}) & (DSC\textcolor{red}{↑})\\
\midrule
SA & 28.336 & 15.56 & 86.30\\
RSA & 18.428 & 10.73 & 88.73 \\
LA & \textbf{15.919} & \textbf{8.61} & 91.16 \\
SA+RSA & 21.622 & 12.60 & 86.89 \\
SA+LA & 20.594 & 11.35 & 90.47 \\
LA+RSA & 17.548 & 9.54 & \textbf{92.09} \\
\bottomrule
\end{tabular}
\end{table}

5) \textbf{Discussion on the actual clinical application}: Considering the LamFormer's superior performance, low Floating Point Operations (FLOPs), and low parameter count, it is highly advantageous for real-world clinical applications. The following are several methods for discussing its practical clinical application:
\par \textbf{Multi segmentation category application}: The LamFormer has demonstrated excellent performance on datasets involving various organ categories, validating its utility for segmentation tasks across different human organs in practical applications. LamFormer can meet the clinical needs of different scenarios without affecting its performance.
\par \textbf{Resource-Efficient Deployment}: Given LamFormer's low computational requirements, the model can be deployed on edge devices or low-power systems, making it accessible in resource-constrained environments such as remote clinics or mobile medical units.
\par \textbf{Interoperability with Existing Systems}: The proposed model can be designed to be interoperable with existing electronic health record systems and medical imaging databases. This integration facilitates the seamless transfer and analysis of patient data, enhancing the overall clinical workflow.
\par In conclusion, the aforementioned discussion has highlighted the potential of the model for practical clinical applications, thereby contributing to enhanced support and service delivery in the field of clinical medicine.

\section{Conlusion}
We propose an enhanced pyramid feature network named LamFormer for fine-grained segmentation of complex multi-organ datasets. This method introduces the Linear Attention Mamba and the Reduced Transformer blocks, aiming to capture long-range dependencies from different-level features. Moreover, we present the Parallel Hierarchical Feature Aggregation module, which integrates multi-scale features from different stages. This module reduces semantic gaps between low- and high-dimensional features and filters redundant information in distinct hierarchical features. LamFormer achieves impressive performance and strong generalization capability on seven diverse and complicated datasets compared to existing methods.
\par Although our method demonstrated excellent segmentation performance on seven medical image datasets, it lacks consideration for datasets from diverse domains. Given the differences in data sources, characteristics, annotation difficulties, and application scenarios across various domains, the current model is unable to effectively conduct cross-modal dataset validation. Future research aims to develop an image segmentation model that is applicable to datasets from different domains. We hope that our research will contribute to the advancement of the medical field.

\section*{References}
\bibliographystyle{ieeetr}
\bibliography{main}

\end{document}